\documentclass[runningheads]{llncs}

\usepackage{eccv}
\usepackage[accsupp]{axessibility} 

\usepackage{eccvabbrv}
\usepackage{gensymb}

\usepackage{graphicx}
\usepackage{booktabs}

\usepackage[accsupp]{axessibility}  

\usepackage{hyperref}

\usepackage{orcidlink}

\begin{document}

\title{Multi-Scale and Multimodal Species Distribution Modeling}


\author{Nina van Tiel\inst{1}\orcidlink{0009-0003-7159-8783} \and
Robin Zbinden\inst{1}\orcidlink{0009-0000-3173-403X} \and
Emanuele Dalsasso\inst{1}\orcidlink{0000-0001-7170-9015}
Benjamin Kellenberger\inst{2}\orcidlink{0000-0002-2902-2014} \and
Loïc Pellissier\inst{3,4}\orcidlink{0000-0002-2289-8259} \and
Devis Tuia\inst{1}\orcidlink{0000-0003-0374-2459}}

\authorrunning{N.~van Tiel et al.}

\institute{Ecole Polytechnique Fédérale de Lausanne, Lausanne, Switzerland \and University College London, London, United Kingdom \and ETH Z\"urich, Zu\"urich, Switzerland \and Swiss Federal Institute for Forest, Snow and Landscape Research, WSL, Birmensdorf, Switzerland}

\maketitle
\begin{abstract}
Species distribution models (SDMs) aim to predict the distribution of species by relating occurrence data with environmental variables. Recent applications of deep learning to SDMs have enabled new avenues, specifically the inclusion of spatial data (environmental rasters, satellite images) as model predictors, allowing the model to consider the spatial context around each species' observations. However, the appropriate spatial extent of the images is not straightforward to determine and may affect the performance of the model, as scale is recognized as an important factor in SDMs.
We develop a modular structure for SDMs that allows us to test the effect of scale in both single- and multi-scale settings. Furthermore, our model enables different scales to be considered for different modalities, using a late fusion approach. Results on the GeoLifeCLEF 2023 benchmark indicate that considering multimodal data and learning multi-scale representations leads to more accurate models. 
\end{abstract}

\section{Introduction}

In the face of the current biodiversity crisis, biodiversity models informed by ever-growing data are crucial to support conservation efforts~\cite{pollock2020protecting}. 
In particular, information about the suitability of species in areas where few observations are recorded enables robust decision-making~\cite{guisan2013predicting}. 
Such information is obtained from species distribution models (SDMs), which relate species occurrence data with environmental variables through statistical methods~\cite{elith2009species,beery2021species,vantiel2023regional}. 

With the increasing availability of species occurrence data, notably through crowd-sourcing ~\cite{joly2016crowdsourcing,van2018inaturalist}, the use of deep learning (DL) has recently been explored in SDMs~\cite{botella2018deep,teng2024satbird,zbinden2024selection}. DL models are of particular interest thanks to their flexibility in terms of architecture and input data types, their scalability, and their ability to model the distributions of many species with a single model~\cite{estopinan2022deep,cole2023spatial, zbinden2023exploring,brun2024multispecies}.

While most SDM approaches use point values of environmental variables as predictors (\emph{i.e.} tabular data), DL facilitates the integration of geospatial information, which allows the model to consider the spatial context surrounding each species observation. Recent works have used convolutional neural networks (CNNs) to integrate patches extracted from rasters of environmental variables~\cite{botella2018deep,deneu2021convolutional}, satellite images~\cite{estopinan2022deep, dollinger2024sat}, or both~\cite{teng2024satbird}. However, the size of the image patches considered by the model is often not justified. 
Studies on non-DL SDMs have shown that scale affects model performance and that the appropriate scale may depend on the species or the type of environment 
\cite{guisan2005predicting,konig2021scale,lu2023scale}. Yet, when handling tabular data, these effects are essentially linked to the resolution of the predictors. 
To the best of our knowledge, no study has focused on the effect of spatial extent\footnote{We use \textit{scale} or \textit{spatial extent} to refer to the size of image patches considered by the model (i.e. its receptive field), and \textit{resolution} to refer to the size of the pixels.} of the image patches on the performance of DL-SDMs. 

Furthermore, little work has gone into considering spatial data with different resolutions in multimodal DL-SDMs. One study used the same patch size for all images, treating them as a single stack but resulting in different spatial extents for each data source~\cite{deneu2021convolutional}. Others have aligned the resolutions of the different images in pre-processing, even though this leads to unnecessarily large models for coarser grain images \cite{teng2024satbird}. Finally, another study used satellite images alongside coarser bioclimatic data considered as tabular data, hence not considering the spatial context for the latter modality~\cite{dollinger2024sat}. 

In this study, we analyze the effects of scale in a modular structure for SDMs based on CNNs. Inspired by works in multi-scale modeling~\cite{chen2017deeplab, reed2023scale} and multi-modal modeling with late fusion~\cite{mac2019presence, dollinger2024sat}, we design a model that can extract features at multiple scales from a single feature map and from multiple modalities with different resolutionsu architecture enables each modality to be considered at its native resolution and at different scales. Using the GeoLifeCLEF 2023 (GLC23) benchmark~\cite{botella2023geolifeclef}, we investigate the effect of spatial extent on model performance in both single- and multi-scale settings, as well as uni- and bi-modal models. Our results indicate superior performance of multi-scale, multimodal approaches. Code to replicate our models can be found on \href{https://github.com/ninavantiel/multi_scale_SDMs}{GitHub}.

\section{Methods}

\subsubsection{Species data}
The GLC23 dataset \cite{botella2023geolifeclef} contains two types of georeferenced plant species observation data. 
It includes 5 million presence-only (PO) observations across Europe. This type of data consists of opportunistic observations in which the non-observation of a species does not confirm its absence. It is widely available but is subject to many biases~\cite{di2021observing}. 
Additionally, presence-absence (PA) data is available for $26\text{k}$ sites in France and Great Britain. PA data is more difficult to obtain as it reflects exhaustive sampling with confirmed species absences. It represents species distribution more accurately but is often not available~\cite{elith2020presence}. 
To reflect the reality of available data for most species, we train our models on PO data and validate them with PA data. In the GLC23 challenge, all teams used PA data for training~\cite{botella2023overview}, thus our results are not comparable to those of the leaderboard.
As only part of the PA data is openly available in GLC23, we use $7,438$ PA sites for validation. The labels for the remaining $22,404$ sites are kept secret for evaluation through the \href{https://www.kaggle.com/competitions/geolifeclef-2023-lifeclef-2023-x-fgvc10}{GLC23 Kaggle page}. We use this second dataset as a test set, even though it only allows us to report the evaluation metric used by the challenge. For training, we keep the PO occurrences for the $2,173$ species in the validation set and merge occurrences recorded at the same location and date, amounting to $2,856,818$ training samples.

\subsubsection{Model predictors}
We use two modalities among those included in GLC23 dataset \cite{botella2023geolifeclef}: $19$ bioclimatic rasters describing temperature and precipitation at a $30$-arc seconds resolution ($\approx 600$ m at $50\degree$ N, the median latitude of occurrence records)~\cite{karger2017climatologies} and images from the Sentinel-2 satellite, providing RGB and a near-infrared (NIR) bands at a $10$-meter resolution. 
All input data is normalized by subtracting the mean and dividing by the standard deviation. From the bioclimatic rasters, patches of various sizes between $1 \times 1$ and $25 \times 25$ pixels are extracted around species occurrence records, corresponding to a ground footprint from $0.6 \text{km} \times 0.6 \text{km} = 0.36 \text{km}^2$, up to $15 \text{km} \times 15 \text{km} = 225 \text{km}^2$. For the satellite data, the dataset provides patches of $128 \times 128$ pixels, centered around the species occurrence records. We extract patches with sizes of $25 \times 25$, $59 \times 59$, and $115 \times 115$ pixels, which correspond to a ground footprint of $0.25 \text{km} \times 0.25 \text{km} = 0.06 \text{km}^2$, $0.59 \text{km} \times 0.59 \text{km} = 0.35 \text{km}^2$, and $1.15 \text{km} \times 1.15 \text{km} = 1.33 \text{km}^2$, respectively. Our models take as input image patches with the size required for the largest spatial extent considered for each modality.

\subsubsection{Model}
\begin{figure}[tb]
    \centering
    \includegraphics[width=\textwidth]{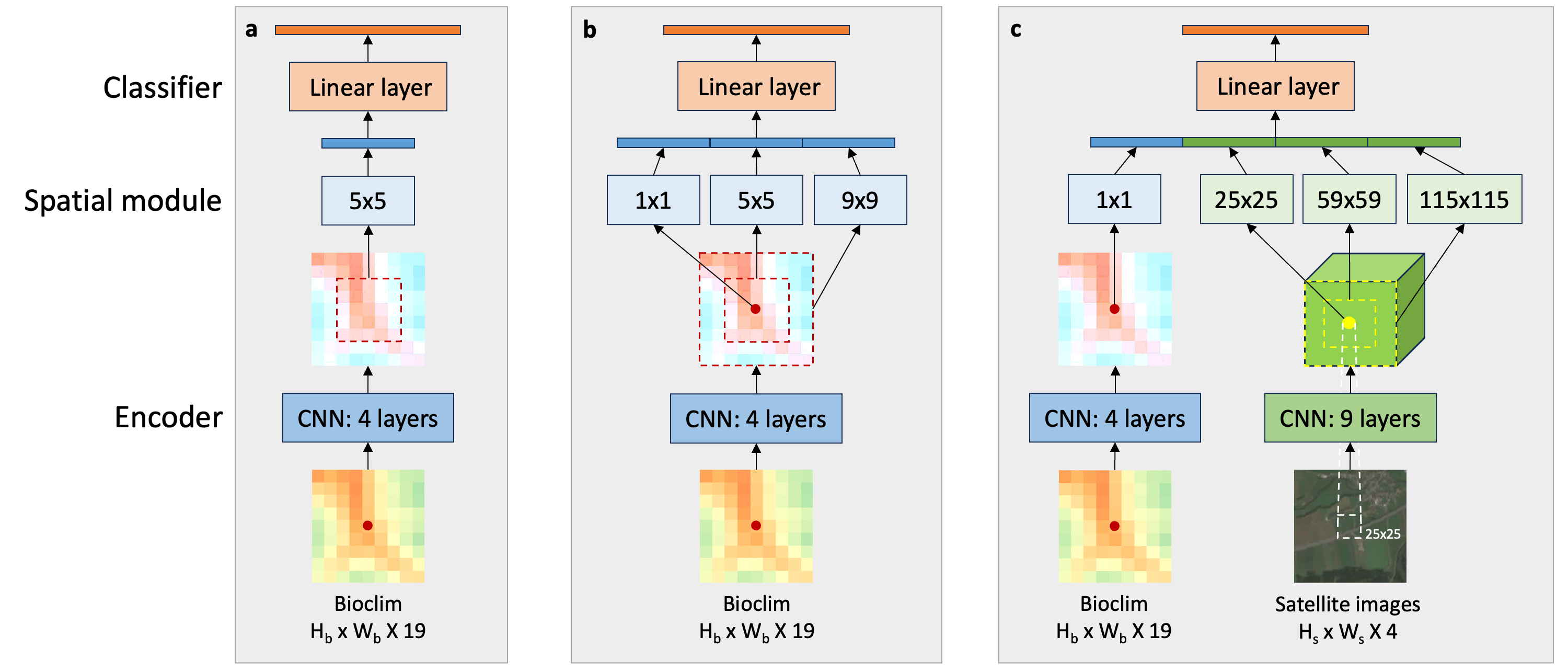}
    \caption{Example of architectures with our modular structure for SDMs. \textbf{a.} Single-scale, unimodal model architecture for bioclimatic variables at scale $(5 \times 5)$. \textbf{b.} Multi-scale unimodal model architecture for bioclimatic variables at scales $(1\times 1)$, $(5 \times 5)$ and $(9\times 9)$. \textbf{c.} Multi-scale multimodal model architecture for bioclimatic variables at scale $(1\times 1)$, and Sentinel-2 satellite images at scales $(25 \times 25)$, $(59\times 59)$ and $(115\times 115)$. The receptive fields after the encoders for bioclimatic variables and satellite image are $1\times 1$ and $25 \times 25$ pixels, respectively.}
    \label{fig:model}
\end{figure}

We propose a model structured in three parts: (1) a common encoder for all scales, (2) a spatial module that can have one or multiple branches for single- or multi-scale models, and (3) a linear classification layer with the same number of output neurons as species, $2,174$ in our case, that is applied to the concatenated outputs of the previous module (Fig.~\ref{fig:model}a,b). The sigmoid function is applied to the output to obtain predictions between $0$ and $1$.
When considering multiple modalities, each one is encoded separately and the spatial module is adapted to the resolution and the scales to be considered for each modality. We use late-fusion and concatenate the 1024-dimensional feature vectors output by each branch of the spatial modules before the final classification layer (Fig.~\ref{fig:model}c). Late-fusion has been shown to work well in single-scale multimodal SDMs~\cite{dollinger2024sat}. Our approach, where modality-specific streams are only fused at the end, allows us to consider each modality at its native resolution and at different scales.

The encoder for bioclimatic variables is composed of four convolutional layers with a kernel size of $1$, keeping the encoder's receptive field at $1$ pixel, or $0.6 \text{km} \times 0.6 \text{km}$. Between each convolution, batch normalization and ReLU are applied. The receptive field of $1$ allows the downstream spatial module to consider a small number of pixels as spatial extent, which is pertinent with the coarse resolution of $30$-arc seconds.
The satellite image encoder consists of nine convolutional layers, corresponding to the first layers up to the second residual block of a ResNet model~\cite{he2016deep}, resulting in a receptive field of $25 \times 25$ pixels or $0.25 \text{km} \times 0.25 \text{km}$.

The spatial module may consist of one or multiple branches, where the number of branches corresponds to the number of scales taken into account. Each branch consists of a series of convolutional and max pooling layers, after which the central pixel of the tensor is extracted to obtain a vector of length $512$. The receptive field of the central pixel corresponds to the extent considered by that branch of the spatial module. Finally, a linear layer with $1,024$ outputs and ReLU are applied to each vector.


We use a weighted loss function for multi-label classification for SDMs~\cite{zbinden2024selection}, which was shown to perform well on the GLC23 dataset~\cite{zbinden2024imbalance}. We consider the records of other species as negatives and do not sample additional pseudo-absences, to avoid the costly operation of downloading new satellite images on the fly. We use a stochastic gradient descent optimizer with a learning rate of $0.01$ and weight decay of $0.0001$. All models are trained end-to-end for $30$ epochs with a batch size of $256$ on NVIDIA A100 with $80$GB video memory.

\subsubsection{Evaluation}
We evaluate our models on the validation set with the area under the receiver operating characteristic curve (AUC). AUC is widely used in SDMs and measures how well the model discriminates presence from absence sites for each species~\cite{valavi2022predictive}. We consider the median AUC across species. 
Additionally, we compute the metric used for the GLC23 challenge~\cite{botella2023overview}, the micro-F1 score, on the validation and test sets. This metric measures the overlap between the predicted and actual set of species, averaged over the sites. While AUC can be computed directly on the probabilistic output of our models, the F1 requires binary predictions. Although species- or model-specific binarization schemes may be used, we chose a fixed binarization threshold of 0.5 to ensure comparability among models and avoid overfitting on the validation data.

\section{Results}

\subsubsection{Unimodal models}

\begin{table}[tb]
  \caption{Model performance and training time for unimodal models considering various scales in single- and multi-scale settings. The median species AUC is computed on the validation data. The micro-F1 score is computed on the validation and test data. The best and second-best scores per column are in bold and underlined, respectively. Scales are indicated in pixels and performance metrics are in \%.}
  \label{tab:main_table}
  \centering
    \begin{minipage}{.45\linewidth}
    \centering
    \label{tab:bioclim_table}
    \begin{tabular}{@{}lllll@{}}
        \toprule
        & \multicolumn{2}{c}{Validation} & Test & \\
        Scales & AUC & F1 & F1 & Runtime\\ 
        \midrule
        1 & \textbf{86.91} & 3.05 & \underline{2.37} & 1.5 hrs\\
        5 & 85.69 & \textbf{3.16} & \underline{2.37} & 2.6 hrs\\
        9 & 83.63 & 2.95 & 2.36 & 2.0 hrs\\
        17 & 83.59 & 2.30 & 2.08 & 3.6 hrs\\
        25 & 83.65 & 1.98 & 1.58 & 9.7 hrs\\\hline
        1,5 & 85.73 & 3.13 & \textbf{2.39} & 2.8 hrs\\
        1,5,9 & 85.37 & \underline{3.14} & 2.30 & 2.9 hrs\\
        1,5,9,17 & \underline{86.28} & 3.05 & 2.35 & 6.8 hrs\\
        1,5,9,17,25 & 85.12 & 3.13 & 2.28 & 17.9 hrs\\
        \bottomrule
    \end{tabular}\\[4pt]
    (a) Bioclimatic variables
    \end{minipage}
    \hspace{0.05\linewidth} 
    \begin{minipage}{.45\linewidth}
    \centering
    \label{tab:sat_table}
    \begin{tabular}{@{}llllll@{}}
        \toprule
        & \multicolumn{2}{c}{Validation} & Test & \\
        Scales & AUC & F1 & F1 & Runtime\\
        \midrule
        25 & 80.41 & 2.76 & 1.69 & 3.9 hrs\\
        59 & \underline{81.38} & 2.88 & 1.98 & 8.0 hrs\\
        115 & 80.85 & 2.98 & 2.00 & 28.7 hrs\\\hline
        25,59 & 80.67 & \underline{3.19} & \underline{2.15} & 9.6 hrs\\
        25,59,115 & \textbf{81.80} & \textbf{3.53} & \textbf{2.25} & 38.6 hrs\\
        \bottomrule
    \end{tabular}\\[4pt]
    (b) Satellite images
    \end{minipage}
\end{table}

First, we train models with only bioclimatic variables or satellite images as predictors. Table~\ref{tab:main_table} shows the performance and training time of single- and multi-scale models with different spatial extents. 

Table~\ref{tab:bioclim_table}a shows that, when considering only bioclimatic variables, small spatial extents obtain the best performance. While the performance decreases with increasing spatial extent for single-scale models, the different multi-scale models recover the performance of the best single-scale models, albeit with longer training time.
We note that the $1 \times 1$ scale obtains the best AUC, and the $5 \times 5$ scale yields the best F1 score on the validation set. Interestingly, the combination of these scales slightly outperforms both single-scale models on the test set, indicating the marginal advantage of a multi-scale approach for this modality.
The differences in performance among these models are relatively small. This may be explained by the high spatial autocorrelation in bioclimatic variables, leading to limited additional information for medium-sized spatial contexts. 
However, considering large spatial extents can be disadvantageous, indicating that, beyond a certain extent, spatial context is not informative and may even introduce spurious correlations for modeling the distributions of plant species. 

In contrast, the results for models with satellite images in Table~\ref{tab:sat_table}b indicate that taking multiple scales into account leads to better performance. While the F1 scores generally increase with model complexity, the AUC does not consistently follow this trend, with no clear relationship between spatial extent and performance among the single-scale models.
This result suggests a larger variability of which scales are most informative, possibly due to the much higher resolution and semantic content of satellite images compared to bioclimatic variables. 
We speculate that different scales may be informative for different species or sites and that the multi-scale architecture may learn which spatial extents are relevant, leading to its superior performance.

\subsubsection{Multimodal models}

\begin{figure}[tb]
    \centering
    \includegraphics[width=\textwidth]{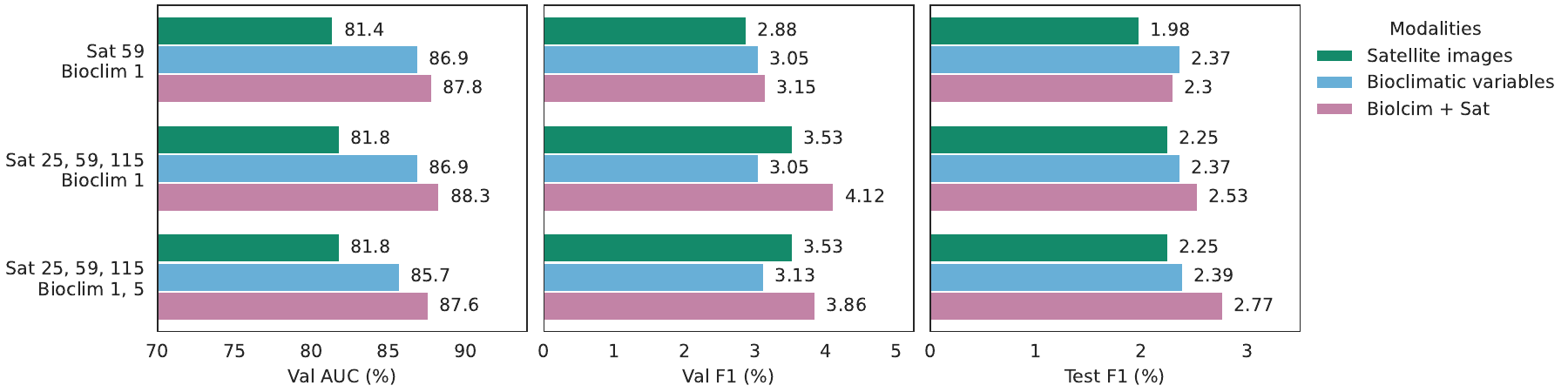}
    \caption{Performance of bimodal and corresponding unimodal models, quantified by their validation median AUC, and test micro-F1 scores.}
    \label{fig:multimodal}
\end{figure}

We compare the performance of bimodal models to their unimodal counterparts (Fig.~\ref{fig:multimodal}). We consider the best scale or combination of scales for each modality. 
Concerning the AUC, we find that the models with bioclimatic variables outperform those with satellite images, and combining both modalities leads to further improvement. 
Regarding the site-wise performance, quantified by the F1 scores, we find that the unimodal models with multiple scales for satellite images outperform the models with bioclimatic variables on the validation set, but the opposite is observed on the test set. The bimodal models perform better than both of their unimodal counterparts, in particular with multi-scale feature extraction for the satellite imagery.  
Furthermore, the inclusion of multiple scales for both bioclimatic variables and satellite images leads to the best F1 score on the test set.
These results confirm the advantage of combining modalities describing the environmental conditions with satellite images~\cite{dollinger2024sat}. Furthermore, they indicate that multi-scale representations for both modalities lead to better species community predictions.
However, such complex models require $20$-fold longer training times than the simplest unimodal models. One may consider whether the performance increase is worth the carbon emissions associated with training these models, estimated at $6.91$ and $0.35$ kg, respectively~\cite{lacoste2019quantifying}.

\subsubsection{Species- and site-level differences in performance}

\begin{figure}[tb]
    \centering
    \includegraphics[width=\textwidth]{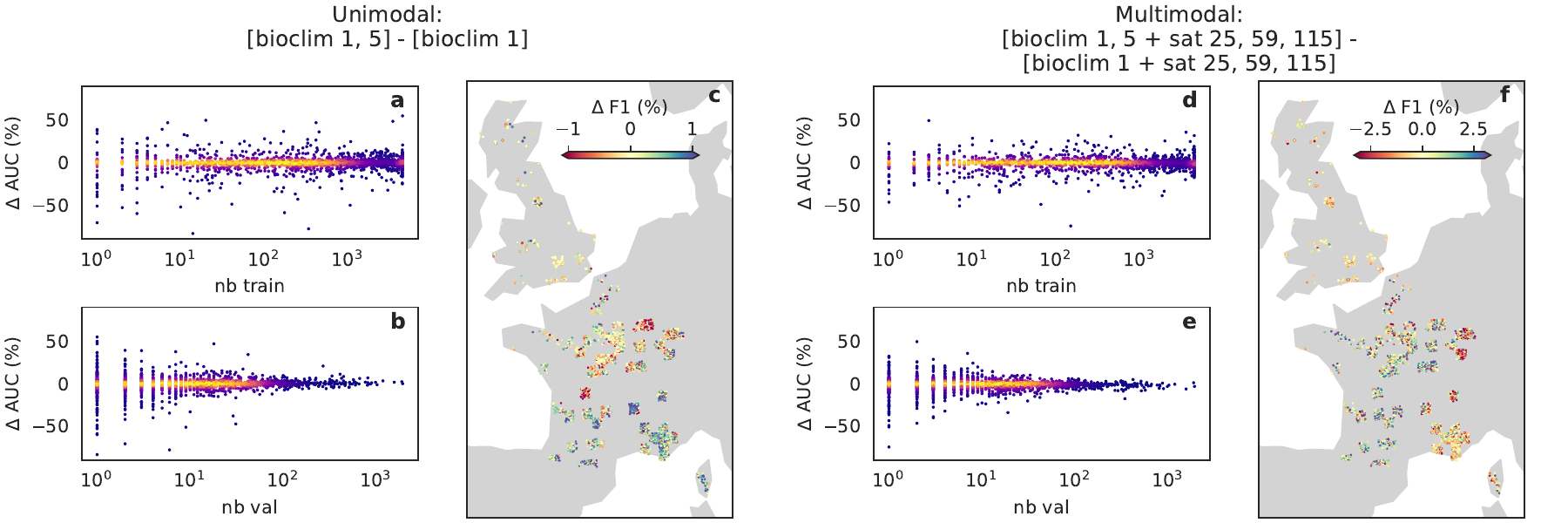}
    \caption{Difference in median AUC ($\Delta$AUC) and micro-F1 ($\Delta$F1) between two unimodal and two bimodal models. Positive $\Delta$ values indicate that models [bioclim 1,5] or [bioclim 1, 5 + sat 25, 59, 115] outperform models [bioclim 1] or [bioclim 1 + sat 25, 59, 115], respectively, and vice-versa for negative values. 
    \textbf{a, b, d, e.} $\Delta$AUC values plotted against the number of occurrences in the training data (nb train) and the validation data (nb val) for $2,173$ species. Colors indicate point density, with higher densities in yellow.
    \textbf{c, f.} $7,348$ validation sites plotted on maps and colored by $\Delta$F1.}
    \label{fig:delta_auc_f1}
\end{figure}

While the differences in performance are sometimes relatively small when aggregated, some species or sites have large differences among models. To illustrate this, we compare two models with bioclimatic variables as predictors and two bimodal models (Fig.~\ref{fig:delta_auc_f1}). These pairs of models have differences in median AUC of $1.2\%$ and $0.7\%$, respectively, and, accordingly, the vast majority of species have a small difference in performance: $90\%$ of species have a difference in AUC of less than $11\%$ between the two pairs of models.
The species with larger differences have few records in the validation data (Fig.~\ref{fig:delta_auc_f1}a,d), but no clear trend can be found with the number of records in the training data (Fig.~\ref{fig:delta_auc_f1}b,e). While this may be explained by the sensitivity of AUC to small sample sizes~\cite{valavi2022predictive}, our results suggest that rare species may be more sensitive to the scale of the predictors.
Mapping the differences in the validation F1 score per site reveals geographical clusters (Fig.~\ref{fig:delta_auc_f1}c,f). These clusters are more locally marked for bioclimatic variables, with a clear preference for a certain model in some regions, despite a median difference in F1 of $0.03\%$. For the bimodal models, considering a single scale for bioclimatic variables generally leads to better site-wise performance, with a median difference in F1 of $-0.23\%$. However, this difference is less marked in certain regions, such as the north and west of France. 
These qualitative results indicate that multi-scale models may be more informative for some species or in some regions, characterized by a specific type of ecosystem. We leave to future work the further investigation of these relationships.

\section{Conclusion}
In this study, we develop a modular structure for SDMs and explore the effect of the scale of spatial predictor variables on the GLC23 dataset for European plant species distributions. 
We find that small scales are most appropriate when considering bioclimatic variables. 
When using satellite images, our multi-scale approach showed a clear benefit in performance.
Combining the best architectures for each modality with a late fusion scheme leads to further increases in performance, indicating the complementarity of both modalities.
Overall, our multi-scale and multimodal model achieved the best performances for both species-wise and site-wise evaluation. 
Our results suggest that the most informative scales may depend on the species or site. Future work may explore these relationships further, and investigate scale-dependencies in other species groups beyond plants.

\section*{Acknowledgements}
The authors acknowledge funding from the deepHSM project with the national funder Swiss National Science Foundation (204057).

%
%
\bibliographystyle{splncs04}
\bibliography{main}
\end{document}